\documentclass[conference]{IEEEtran}
\IEEEoverridecommandlockouts
\usepackage{cite}
\usepackage{amsmath,amssymb,amsfonts}
\usepackage{algorithmic}
\usepackage{graphicx}
\usepackage{textcomp}
\usepackage{xcolor}
\usepackage{multirow}
\def\BibTeX{{\rm B\kern-.05em{\sc i\kern-.025em b}\kern-.08em
    T\kern-.1667em\lower.7ex\hbox{E}\kern-.125emX}}
\begin{document}

\title{CT-PatchTST: Channel-Time Patch Time-Series Transformer for
Long-Term Renewable Energy Forecasting}

% \author{\IEEEauthorblockN{1\textsuperscript{st} Given Name Surname}
% \IEEEauthorblockA{\textit{dept. name of organization (of Aff.)} \\
% \textit{name of organization (of Aff.)}\\
% City, Country \\
% email address or ORCID}
% \and
% \IEEEauthorblockN{2\textsuperscript{nd} Given Name Surname}
% \IEEEauthorblockA{\textit{dept. name of organization (of Aff.)} \\
% \textit{name of organization (of Aff.)}\\
% City, Country \\
% email address or ORCID}
% \and
% \IEEEauthorblockN{3\textsuperscript{rd} Given Name Surname}
% \IEEEauthorblockA{\textit{dept. name of organization (of Aff.)} \\
% \textit{name of organization (of Aff.)}\\
% City, Country \\
% email address or ORCID}
% \and
% \IEEEauthorblockN{4\textsuperscript{th} Given Name Surname}
% \IEEEauthorblockA{\textit{dept. name of organization (of Aff.)} \\
% \textit{name of organization (of Aff.)}\\
% City, Country \\
% email address or ORCID}
% \and
% \IEEEauthorblockN{5\textsuperscript{th} Given Name Surname}
% \IEEEauthorblockA{\textit{dept. name of organization (of Aff.)} \\
% \textit{name of organization (of Aff.)}\\
% City, Country \\
% email address or ORCID}
% \and
% \IEEEauthorblockN{6\textsuperscript{th} Given Name Surname}
% \IEEEauthorblockA{\textit{dept. name of organization (of Aff.)} \\
% \textit{name of organization (of Aff.)}\\
% City, Country \\
% email address or ORCID}
% }

\author{
% Left column authors

\IEEEauthorblockN{1\textsuperscript{st} Kuan Lu\textsuperscript{\dag *}}
\IEEEauthorblockA{\textit{School of Electrical and Computer Engineering} \\
\textit{Cornell University} \\
Ithaca, NY, USA \\
kl649@cornell.edu \\
ORCID: 0009-0003-5744-9247}

\vspace{+1em}

\IEEEauthorblockN{2\textsuperscript{nd} Yuxiao Li}
\IEEEauthorblockA{\textit{Department of Electrical and Computer Engineering} \\
\textit{Northeastern University} \\
Boston, MA, USA \\
li.yuxiao@northeastern.edu \\
ORCID: 0009-0002-9813-5925}

\vspace{+1em}

\IEEEauthorblockN{4\textsuperscript{th} Zhenrui Chen}
\IEEEauthorblockA{\textit{Fu Foundation School of Engineering and Applied Science} \\
\textit{Columbia University in the City of New York} \\
New York, NY, USA \\
zc2569@columbia.edu \\
ORCID: 0009-0007-6909-3638}

\and

\IEEEauthorblockN{1\textsuperscript{st} Menghao Huo\textsuperscript{\dag}}
\IEEEauthorblockA{\textit{School of Engineering} \\
\textit{Santa Clara University} \\
Santa Clara, CA, USA \\
menghao.huo@alumni.scu.edu \\
ORCID: 0009-0000-0076-5343}

\vspace{+1em}

% Right column authors
\IEEEauthorblockN{3\textsuperscript{rd} Qiang Zhu}
\IEEEauthorblockA{\textit{Department of Mechanical and Aerospace Engineering} \\
\textit{University of Houston} \\
Houston, TX, USA \\
qzhu11@uh.edu \\
ORCID: 0009-0002-0981-0635}

\thanks{*Corresponding author: Kuan Lu (kl649@cornell.edu)}
\thanks{\textsuperscript{\dag}Kuan Lu and Menghao Huo contributed equally to this work.}
}

\maketitle

\begin{abstract}
Accurate forecasting of renewable energy generation is fundamental to enhancing the dynamic performance of modern power grids, especially under high renewable penetration. This paper presents Channel-Time Patch Time-Series Transformer (CT-PatchTST), a novel deep learning model designed to provide long-term, high-fidelity forecasts of wind and solar power. Unlike conventional time-series models, CT-PatchTST captures both temporal dependencies and inter-channel correlations—features that are critical for effective energy storage planning, control, and dispatch. Reliable forecasting enables proactive deployment of energy storage systems (ESSs), helping to mitigate uncertainties in renewable output, reduce system response time, and optimize storage operation based on location-specific flow and voltage conditions. Evaluated on real-world datasets from Denmark’s offshore wind, onshore wind, and solar generation, CT-PatchTST outperforms existing methods in both accuracy and robustness. By enabling predictive, data-driven coordination of ESSs across integrated source–grid–load–storage systems, this work contributes to the design of more stable, responsive, and cost-efficient power networks.
\end{abstract}


\section{Introduction}
Renewable energy has become a cornerstone of modern power systems, driven by the urgent need for sustainable and low-carbon energy solutions. However, its inherent variability poses serious challenges to real-time power balancing and grid stability. Energy storage systems (ESSs) have emerged as key enablers to mitigate such volatility, by shifting renewable output temporally and enabling dynamic response to grid conditions. Accurate forecasting of wind and solar generation is thus essential not only for day-ahead scheduling but also for the optimal planning, control, and dispatch of ESSs within integrated source–grid–load–storage frameworks. This paper introduces CT-PatchTST, a novel transformer-based forecasting model designed to meet these practical demands by enhancing the temporal and structural understanding of multivariate renewable datasets.

Forecasting power generation is essential for maintaining a stable balance between energy supply and demand, enhancing the efficiency of energy storage systems, and mitigating risks of grid instability. Accurate predictions enable energy providers, grid operators, and policymakers to plan energy distribution, storage, and operational strategies more effectively. However, renewable energy forecasting poses unique challenges due to its dependence on external and often unpredictable natural factors \cite{ouyang2017model}. For instance, wind power generation is influenced by variables such as wind speed, direction, and atmospheric temperature, while solar power generation depends on sunlight intensity, cloud cover, and seasonal variations. These factors introduce nonlinear and dynamic patterns in power generation, further compounded by diurnal and seasonal cycles, making reliable forecasting a complex yet vital task.

Traditional methods for renewable energy forecasting, such as statistical models and physical simulations, have provided a strong foundation for the field but face significant limitations in handling the intricate temporal dependencies and nonlinear relationships inherent in renewable energy data. Statistical models like ARIMA, while effective for short-term forecasting, often struggle to capture long-term dependencies and adapt to abrupt changes in data patterns \cite{liu2016online,adebiyi2014comparison}. On the other hand, physical simulations require extensive domain-specific knowledge and computational resources, limiting their scalability in dynamic real-world scenarios. These challenges underscore the urgent need for advanced and flexible approaches to address the complexities of renewable energy forecasting.

Recent advances in machine learning, particularly deep learning, have provided a variety of powerful tools to address complex and data-intensive challenges. Traditional approaches such as filtering-based methods and support vector machines remain valuable in many scenarios \cite{wu2012extended, joo2015time, han2012real}. In parallel, deep learning has been increasingly applied in areas such as infrastructure analysis, image-based prediction, and sequential decision-making \cite{dan2024multiple, dan2024image, lu2023deep}. Representative architectures, such as Convolutional Neural Networks \cite{li2021survey}, Recurrent Neural Networks \cite{grossberg2013recurrent}, and Transformers \cite{ashish2017attention}, have significantly improved the capacity to model complex spatial and temporal relationships \cite{li2025revolutionizing, zhang2024enhancing,deng20222event,li2025ddunet, ying2020sichuan}. The emergence of large language models has further expanded the capabilities of learning systems, enabling new possibilities in knowledge extraction, interpretation, and task automation \cite{behari2024decision,sehanobish2023scalable,zhao2023towards,tao2024robustness,du2025zero, zhang2025ratt}. These developments are underpinned by ongoing theoretical progress in statistical estimation and optimization, which continue to enhance the generalization and scalability of learning-based models \cite{zhao2022analysis, zhao2024minimax}. Among these methods, the Patch Time-Series Transformer (PatchTST) \cite{nie2022time} has demonstrated remarkable effectiveness in modeling temporal dependencies in multivariate time-series data. However, the original PatchTST employs a channel-independent approach, which, while effective for managing data granularity, often neglects critical inter-channel relationships. This limitation can result in the loss of valuable contextual information essential for accurate forecasting \cite{han2024mcformer, wang2023dance}.To address these shortcomings, we propose an enhanced model, the Channel-Time Patch Time-Series Transformer (CT-PatchTST). Fig. 1 provides an overview of the model architecture. CT-PatchTST builds upon the strengths of PatchTST by introducing channel attention mechanisms that effectively capture and integrate inter-channel dependencies while retaining the benefits of channel-independent modeling.

\begin{figure*}
    \centering
    \includegraphics[width=\textwidth, keepaspectratio]{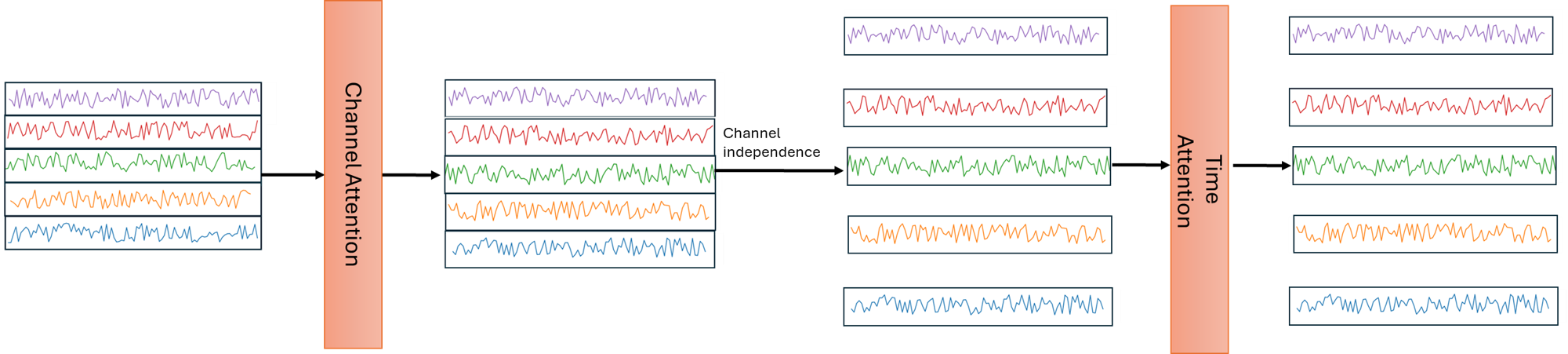}
    \caption{Overview of CT-PatchTST model. After the multivariate time series is processed through channel attention, the inter-channel relationships are learned, resulting in the generation of a new multivariate time series. This transformed series is then passed through time attention, where each channel is treated independently to capture temporal dependencies. Finally, the prediction results are produced based on the integrated outputs of the channel and time attention mechanisms.}
    \label{fig:enter-label1}
\end{figure*}

The key contributions of this paper are as follows:
\begin{itemize}
    \item We propose CT-PatchTST, an enhanced version of the Patch Time-Series Transformer, which incorporates a dual-attention mechanism combining channel attention and time attention. This design effectively captures inter-channel dependencies that were overlooked in the original PatchTST while maintaining the advantages of channel-independent modeling.
    \item CT-PatchTST simultaneously integrates channel and temporal information, enabling the model to process complex multivariate time-series data with greater accuracy. This novel approach significantly improves forecasting performance for renewable energy systems, addressing the challenges posed by dynamic and nonlinear patterns in wind and solar power generation.
    \item Through extensive experiments on real-world renewable energy datasets, including offshore wind power, onshore wind power, and solar power generation, CT-PatchTST demonstrates superior performance compared to existing state-of-the-art models, highlighting its robustness and reliability across diverse forecasting scenarios.
\end{itemize}

CT-PatchTST is specifically designed to better leverage the interplay between temporal and inter-channel information, enabling it to effectively model the intricate dynamics of renewable energy systems. By rigorously analyzing and comparing the performance of CT-PatchTST with the baseline PatchTST and other models, this study establishes the model's superiority in renewable energy forecasting tasks. The findings of this research contribute to improving the predictability and reliability of wind and solar power systems, supporting their seamless integration into energy grids, and advancing the broader adoption of renewable energy technologies. This work represents a significant step toward building smarter, more sustainable energy systems capable of meeting the growing global demand for clean energy.

\section{Related Work}
Time-series forecasting has long been a critical area of study, evolving from traditional statistical models to sophisticated deep learning techniques. Early approaches, such as the AutoRegressive Integrated Moving Average (ARIMA) model, laid the groundwork for this field \cite{foley2012current}. ARIMA's design effectively handles stationary and non-stationary data by incorporating autoregressive, moving average, and differencing components. While powerful for short-term predictions, ARIMA falls short in addressing long-term dependencies and complex non-linear relationships.

Exponential smoothing methods, like the Holt-Winters model, have also been extensively used for capturing trends and seasonality. By assigning diminishing weights to older observations, these models perform well for data with seasonal variations. However, their inability to model multivariate and non-linear patterns limits their application. Gaussian Process Regression (GPR), a probabilistic framework, offers another perspective by modeling time series as distributions over functions, providing both predictions and uncertainty estimates. However, GPR struggles to scale efficiently with large datasets. Related probabilistic models like Naive Bayes and Bayesian Networks have also been used in fields such as vehicle insurance, offering high accuracy and interpretability \cite{wang2024novel}.

To overcome the limitations of traditional methods, deep learning models have emerged as powerful tools for time-series forecasting. Among these, Long Short-Term Memory (LSTM) networks are particularly notable for their ability to model long-term dependencies. By utilizing a cell state to retain relevant information and gate mechanisms to regulate data flow, LSTMs effectively address the vanishing gradient problem \cite{hochreiter1997long}. 

Building on the advancements made by LSTMs, Seq2Seq models have brought further innovation to time-series forecasting. Initially developed for natural language processing tasks, Seq2Seq models employ an encoder-decoder structure where the encoder compresses the input sequence into a context vector, and the decoder generates predictions based on this representation. This architecture enables Seq2Seq models to handle variable-length input and output sequences, making them suitable for time-series prediction tasks with diverse temporal dynamics \cite{sutskever2014sequence}. However, their reliance on fixed-length context vectors can result in information bottlenecks, particularly for long sequences. The introduction of attention mechanisms has helped mitigate this issue by enabling the model to focus on the most relevant parts of the input sequence dynamically. Despite their computational intensity and scaling challenges, Seq2Seq models have demonstrated notable success in capturing complex temporal dependencies, paving the way for more advanced architectures.

Recently, the introduction of Transformer-based architectures has revolutionized time-series forecasting \cite{wang2024timexer, liu2023itransformer, liu2022pyraformer, zhou2021informer, zhang2023crossformer}, with the Patch Time-Series Transformer (PatchTST) emerging as a notable example. PatchTST segments input sequences into non-overlapping patches, reducing sequence length and enabling efficient parallel processing while maintaining temporal granularity. Its self-attention mechanism dynamically assesses the importance of various time steps, making it highly effective in capturing both local and global temporal patterns. This patch-based approach not only enhances scalability but also improves model generalization across diverse datasets. Compared to Seq2Seq models, although it requires careful hyperparameter tuning and considerable computational resources during training, PatchTST has demonstrated superior performance in long-horizon forecasting tasks. 

\section{Method}
\subsection{Problem Definition}
The goal is to predict multivariate time series based on historical data. Specifically, a look-back window 
\(\{\mathbf{x}_1, \mathbf{x}_2, \ldots, \mathbf{x}_L\} \in \mathbb{R}^{L \times M}\), representing the past \(L\) time steps, is used as input to forecast the future sequence of \(T\) time steps, given by 
\(\{\mathbf{x}_{L+1}, \mathbf{x}_{L+2}, \ldots, \mathbf{x}_{L+T}\} \in \mathbb{R}^{T \times M}\).

Here, each observation \(\mathbf{x}_L\) is an \(M\)-dimensional vector containing multivariate features at time step \(L\). The objective is to learn a mapping function that captures temporal dependencies and relationships among the multivariate features in the look-back window, enabling accurate prediction of the target sequence. This problem is particularly challenging due to the need to handle complex temporal patterns, multivariate interactions, and potential non-linearities in the data. Based on the original PatchTST, we proposed CT-PatchTST to fuse the information between the channels of multivariate time series.The architecture of CT-PatchTST is shown in Fig. 2.

\begin{figure*}
    \centering
    \includegraphics[width=\textwidth]{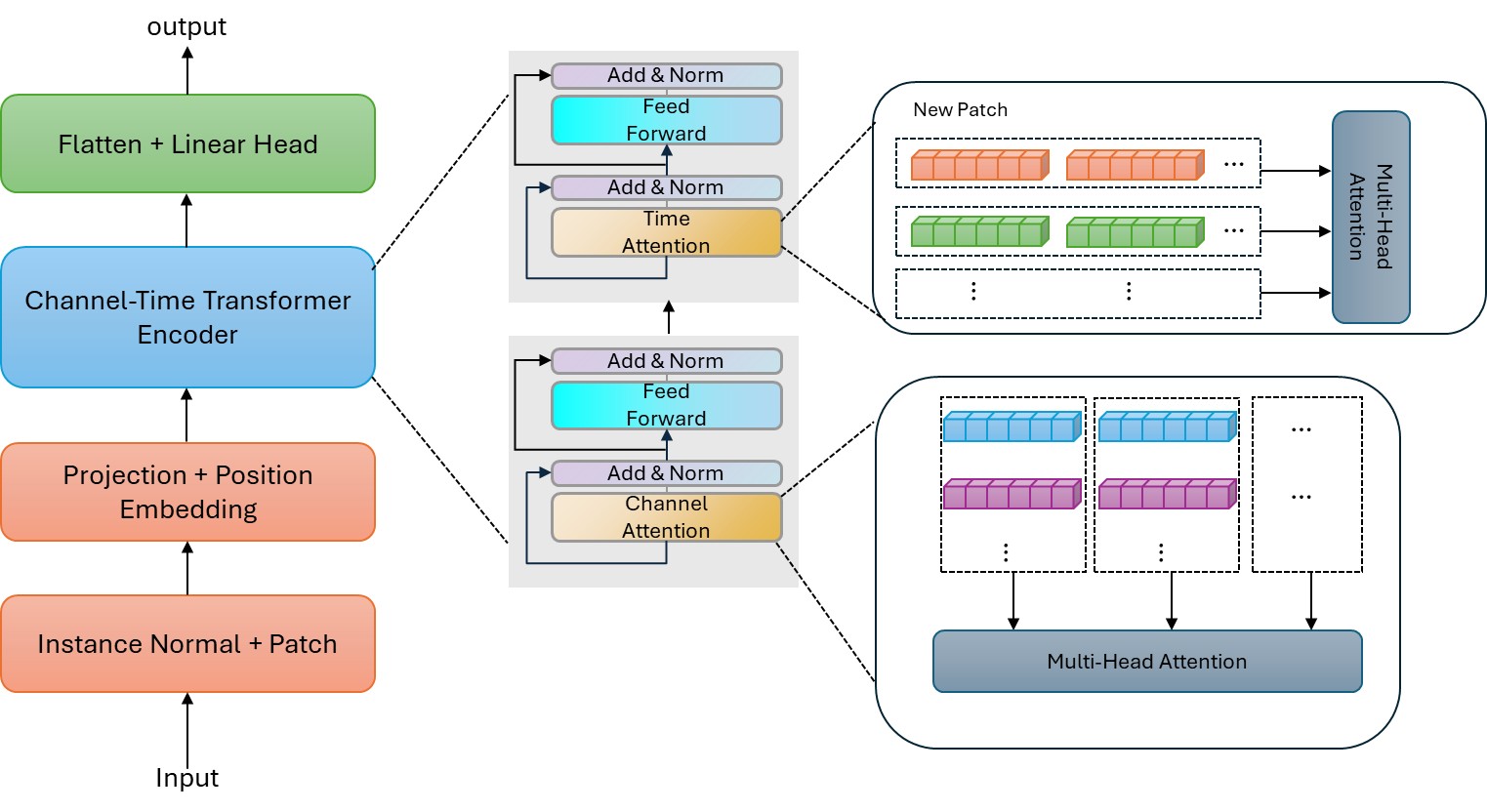}
    \caption{Schematic architecture of the CT-PatchTST. After the multivariate time series is processed through channel attention, inter-channel relationships are learned, producing a new multivariate series. This transformed series is then passed through time attention, where temporal dependencies are captured independently for each channel. The final prediction results are obtained by integrating the outputs of both attention mechanisms.}
    \label{fig:ct-patchtst}
\end{figure*}

\subsection{Reversible Instance Normalization}

Reversible Instance Normalization (RevIN) is designed to mitigate the distribution shift between training and testing data in time-series applications. We apply this technique to normalize the raw data before turning it into patches. This technique operates on each time-series instance $x^{(i)}$, where $x^{(i)}$ represents the sequence of values for the $i$-th feature or channel in the dataset, with $i = 1, 2, \dots, M$. Each $x^{(i)}$ consists of a series of temporal values $[x^{(i)}_1, x^{(i)}_2, \dots, x^{(i)}_L]$.

During normalization, the mean $\text{Mean}(x^{(i)})$ and variance $\text{Var}(x^{(i)})$ are computed over the temporal dimension for the $i$-th feature. The normalized sequence $\tilde{x}^{(i)}$ is then calculated as:
\begin{equation}
\tilde{x}^{(i)} = \frac{x^{(i)} - \text{Mean}(x^{(i)})}{\sqrt{\text{Var}(x^{(i)}) + \epsilon}},
\end{equation}
where $\epsilon$ is a small constant added for numerical stability.

After the normalized sequence $\tilde{x}^{(i)}$ is processed by the model, the output $\tilde{y}^{(i)}$ is denormalized to restore the original data distribution. This is achieved by reapplying the previously computed mean and variance:
\begin{equation}
y^{(i)} = \tilde{y}^{(i)} \cdot \sqrt{\text{Var}(x^{(i)}) + \epsilon} + \text{Mean}(x^{(i)}),
\end{equation}
where $y^{(i)}$ is the restored sequence, ensuring that the non-stationary components of the original data are preserved. 

\subsection{Patch and Projection}
In the study of time prediction problems, using patches as model inputs has become increasingly common, replacing the traditional approach of using individual time points as inputs. This method enhances the semantic information within the input, allowing the model to better capture the relationships between preceding and subsequent time points. As a result, the model's ability to handle time prediction tasks is significantly improved.Therefore, the normalized input time series is first divided into patches to form a sequence of patches. Each patch is then processed through a projection layer (MLP) for dimensional expansion, which enhances the model's ability to fit the original data. Mathematically, this process can be expressed as:
\[
x_p^{(i)} = \text{Projection}(\text{Patch}(\tilde{x}^{(i)}))
\]

Here, \( x_p^{(i)} \in \mathbb{R}^{P \times N} \), where \( P \) represents the dimensionality after the projection, and \( N \) is the total number of patches. The number of patches is calculated as \( N = \left\lfloor \frac{L - P}{S} \right\rfloor + 2 \), where \( L \) is the total length of the input sequence, \( P \) is the patch length, and \( S \) is the stride length. Fig. 3 illustrates the process of patching and projection.
\begin{figure}[htbp]
    \centering
    \includegraphics[width=1\linewidth]{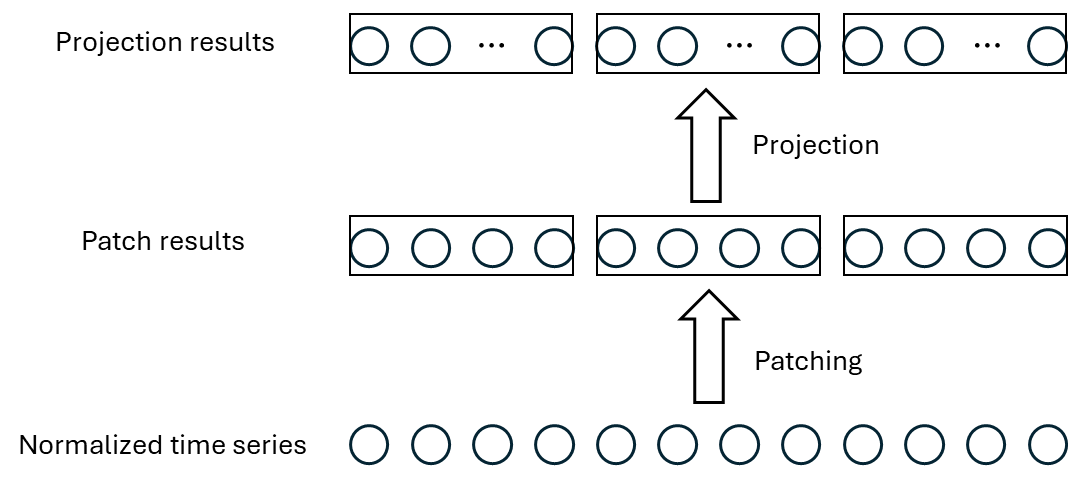}
    \caption{The schematic of patch and projection. Using patch to expand the input semantic information, and using projection to expand the input dimension to improve the fitting ability of the model}
    \label{fig:enter-label3}
\end{figure}
\subsection{Channel-Time Transformer Encoder}
We employed the Channel-Time Transformer encoder to transform the observed multichannel time series into latent representations. The Channel-Time Transformer encoder is a modified version of the vanilla Transformer encoder, incorporating two key mechanisms: channel attention and time attention. These mechanisms work together to model both inter-channel and temporal dependencies in multichannel time series.

After patching, the original input is mapped to $x_p$, where \( x_p\in \mathbb{R}^{P \times N \times M} \), a learnable positional embedding \( \mathbf{W}_{\text{pos}} \in \mathbb{R}^{P \times N} \) is added to retain the temporal order of the sequence. The positional embedding is added along the temporal dimension for each channel:

\[
\mathbf{X}_d = x_p + \mathbf{W}_{\text{pos}},
\]

where \( \mathbf{X}_d \in \mathbb{R}^{P \times N \times M} \). The \( \mathbf{X}_d\) will serve as the input to the subsequent attention mechanisms.

The first step is channel attention, which captures inter-channel dependencies. For each attention head \( h \), the query, key, and value matrices are calculated for each time step \( n = 1, \dots, N \):

\[
\begin{aligned}
\mathbf{Q}_h^{\text{channel},n} = \mathbf{X}_d^{n,T} \mathbf{W}_h^Q, \\
\mathbf{K}_h^{\text{channel},n} = \mathbf{X}_d^{n,T} \mathbf{W}_h^K, \\
\mathbf{V}_h^{\text{channel},n} = \mathbf{X}_d^{n,T} \mathbf{W}_h^V,
\end{aligned}
\]

where \( \mathbf{X}_d^n \in \mathbb{R}^{P \times M} \) represents the features at time step \( n \), and \( \mathbf{W}_h^Q, \mathbf{W}_h^K, \mathbf{W}_h^V \in \mathbb{R}^{P \times d_k} \) are trainable parameters. Note that the attention operates along the channel dimension \( M \). The channel attention output is computed as:

\[
\mathbf{O}_h^{\text{channel},n} = \text{Softmax}\left(\frac{\mathbf{Q}_h^{\text{channel},n} (\mathbf{K}_h^{\text{channel},n})^T}{\sqrt{d_k}}\right) \mathbf{V}_h^{\text{channel},n}.
\]

The outputs from all heads are concatenated and linearly projected, producing the representation \( \mathbf{Z}_{\text{channel}} \in \mathbb{R}^{P \times N \times M} \).

Next, time attention is applied to model temporal dependencies for each channel. Using \( \mathbf{Z}_{\text{channel}} \) as input, the query, key, and value matrices are computed for each channel \( m = 1, \dots, M \):

\[
\begin{aligned}
\mathbf{Q}_h^{\text{time},m} = \mathbf{Z}_{\text{channel}}^{m,T} \mathbf{W}_h^{Q,\text{time}}, \\
\mathbf{K}_h^{\text{time},m} = \mathbf{Z}_{\text{channel}}^{m,T} \mathbf{W}_h^{K,\text{time}}, \\
\mathbf{V}_h^{\text{time},m} = \mathbf{Z}_{\text{channel}}^{m,T} \mathbf{W}_h^{V,\text{time}},
\end{aligned}
\]

where \( \mathbf{Z}_{\text{channel}}^m \in \mathbb{R}^{P \times N} \) represents the features for channel \( m \), and the trainable parameters
\( \mathbf{W}_h^{Q,\text{time}}, \mathbf{W}_h^{K,\text{time}}, \mathbf{W}_h^{V,\text{time}} \in \mathbb{R}^{P \times d_k} \) are used to compute the time attention output.

\[
\mathbf{O}_h^{\text{time},m} = \text{Softmax}\left(\frac{\mathbf{Q}_h^{\text{time},m} (\mathbf{K}_h^{\text{time},m})^T}{\sqrt{d_k}}\right) \mathbf{V}_h^{\text{time},m}.
\]

The outputs from all heads are concatenated and linearly projected, resulting in \( \mathbf{Z}_{\text{time}} \in \mathbb{R}^{P \times N \times M} \).

Each attention layer is followed by a feed-forward layer to enhance the representation within the encoder module. Residual connections and layer normalization are applied after both the attention mechanism and the feed-forward layer to facilitate better integration and stability of the model.

\subsection{Loss Function}

To evaluate the performance of the model, we employ two loss functions:

\begin{itemize}
    \item \textbf{Mean Absolute Error (MAE)}: This loss function calculates the average absolute differences between predicted values $y^{(i)}$ and ground truth values $z^{(i)}$ for all $i = 1, 2, \dots, M$:
    \begin{equation}
    \text{MAE} = \frac{1}{M T} \sum_{i=1}^M \sum_{t=L+1}^T \left| y^{(i)}_t - z^{(i)}_t \right|.
    \end{equation}

    \item \textbf{Mean Squared Error (MSE)}: This loss function calculates the average squared differences between predicted values $y^{(i)}$ and ground truth values $z^{(i)}$:
    \begin{equation}
    \text{MSE} = \frac{1}{M T} \sum_{i=1}^M \sum_{t=L+1}^T \left( y^{(i)}_t - z^{(i)}_t \right)^2.
    \end{equation}
\end{itemize}

Both MAE and MSE are utilized as loss functions to evaluate and optimize the model, where MAE emphasizes robustness to outliers, and MSE penalizes larger deviations more heavily.

\section{Experiment}
\subsection{Dataset}
We evaluate our model using two datasets that comprise measured time series data of renewable energy production, electricity consumption, and energy exchange with neighboring countries and continents, all recorded at an hourly resolution \cite{olsen2019hourly}. The datasets span the period from 2014 to 2019 and contain a total of 499,993 entries. Specifically, we focus on predicting three key variables: 'OffshoreWindPower,' 'OnshoreWindPower,' and 'SolarPowerProd.' These variables provide valuable insights into renewable energy production patterns, making the dataset a robust benchmark for assessing the performance of our model in multivariate time series forecasting tasks.

\subsection{baseline}
Deep learning models have made significant advances in time series forecasting, consistently outperforming traditional methods across diverse tasks. To validate our proposed methodology, we evaluated a selection of state-of-the-art (SOTA) multivariate forecasting approaches. Transformer-based architectures, recognized for their outstanding capabilities in capturing temporal patterns, were a primary focus, including the well-known PatchTST model. Additionally, GRU-based Seq2Seq models, celebrated for their efficiency in extracting multivariate features, were incorporated to ensure a comprehensive comparison.

\subsection{forecasting and baseline comparison}
\textbf{Experimental Settings}
We adopted the experimental settings of PatchTST, configuring the look-back window lengths \(L\) for all models to 336. The prediction window lengths \( T \) were set to \{96, 192, 336, 720\}.

\textbf{Model Variants}
We propose two versions of CT-PatchTST, as shown in Table 1. CT-PatchTST-512 indicates that its look-back window length \( L \) is 512, while CT-PatchTST corresponds to a look-back window length \( L \) of 336. For both models, the patch length \( P \) is set to 16, and the stride \( S \) is defined as \( S = P / 2 \). CT-PatchTST is used as a fair comparison with other models, whereas CT-PatchTST-512 demonstrates the potential for achieving better results. Additional model parameters are provided in Table I.

\begin{table}
\centering
\caption{Hyperparameters of the CT-PatchTST model}
\label{tab:hyperparameters4}
\resizebox{\linewidth}{!}{%
\begin{tabular}{l|c}
\hline
\textbf{Hyperparameter}                                & \textbf{Value} \\ \hline
Number of channel-time encoders                        & 4              \\ 
Head of channel attention                              & 1              \\ 
Head of time attention                                 & 16             \\ 
Dimension of latent space of channel-time encoder      & 256            \\ 
Dimension of latent space of feed-forward network      & 512            \\ 
Batch size                                             & 128            \\ 
Learning rate                                          & 0.001          \\ 
Epochs                                                 & 50             \\ \hline
\end{tabular}%
}
\end{table}

\textbf{Results}
As shown in Table II, our model achieves the best performance compared to other methods. This result shows that applying attention to the channel data before processing the multivariate time series helps the model better learn how the different channels relate to each other. By focusing on the most important parts of the input data early on, the model can build a stronger understanding of the overall structure. This improved understanding leads to more accurate and reliable predictions.

\begin{table*}[!t]
\caption{\textbf{Full Results on the Multivariate Forecasting Task for DK1 (Western Denmark) and DK2 (Eastern Denmark)}. Look-back window is 336 for all. Forecasting horizon $h \in \{96, 192, 336, 720\}$. Best results are in \textbf{bold}, second-best are \underline{underlined}.}
\label{tab:comparison5}
\setlength{\tabcolsep}{2pt}
\renewcommand{\arraystretch}{1.15}
\centering
\begin{tabular}{|c|c|cc|cc|cc|cc|cc|cc|cc|cc|}
\hline
\textbf{Dataset} & \textbf{Horizon (h)} & \multicolumn{2}{c|}{\textbf{CT-PatchTST-512}} & \multicolumn{2}{c|}{\textbf{CT-PatchTST}} & \multicolumn{2}{c|}{\textbf{PatchTST~\cite{nie2022time}}} & \multicolumn{2}{c|}{\textbf{Seq2Seq~\cite{sutskever2014sequence}}} & \multicolumn{2}{c|}{\textbf{iTransformer~\cite{liu2023itransformer}}} & \multicolumn{2}{c|}{\textbf{PAttn~\cite{tan2024language}}} & \multicolumn{2}{c|}{\textbf{TimeXer~\cite{wang2024timexer}}} & \multicolumn{2}{c|}{\textbf{TSMixer~\cite{chen2023tsmixer}}} \\
\hline
 & \textbf{Metric} & \textbf{MSE} & \textbf{MAE} & \textbf{MSE} & \textbf{MAE} & \textbf{MSE} & \textbf{MAE} & \textbf{MSE} & \textbf{MAE} & \textbf{MSE} & \textbf{MAE} & \textbf{MSE} & \textbf{MAE} & \textbf{MSE} & \textbf{MAE} & \textbf{MSE} & \textbf{MAE} \\
\hline
\multirow{4}{*}{\textbf{DK1}}
& 96  & \textbf{0.0060} & \underline{0.0515} & \underline{0.0061} & \textbf{0.0504} & 0.0066 & 0.0540 & 0.0187 & 0.0561 & 0.0070 & 0.0549 & 0.0203 & 0.0893 & 0.0115 & 0.1004 & 0.0121 & 0.0694 \\
& 192 & \textbf{0.0060} & \textbf{0.0505} & \underline{0.0063} & \underline{0.0502} & 0.0070 & 0.0548 & 0.0208 & 0.1190 & 0.0114 & 0.0767 & 0.0225 & 0.0938 & 0.0118 & 0.1007 & 0.0238 & 0.1098 \\
& 336 & \textbf{0.0061} & \textbf{0.0509} & \underline{0.0071} & \underline{0.0532} & 0.0087 & 0.0595 & 0.1573 & 0.2270 & 0.0172 & 0.1260 & 0.0274 & 0.1019 & 0.0147 & 0.1091 & 0.0338 & 0.1354 \\
& 720 & \textbf{0.0077} & \textbf{0.0579} & \underline{0.0093} & \underline{0.0644} & 0.0114 & 0.0683 & 0.1940 & 0.3510 & 0.0291 & 0.1730 & 0.0398 & 0.1198 & 0.0189 & 0.1109 & 0.0572 & 0.1656 \\
\hline
\multirow{4}{*}{\textbf{DK2}}
& 96  & \underline{0.0079} & \textbf{0.0598} & 0.0081 & \textbf{0.0591} & 0.0089 & 0.0634 & 0.0152 & 0.0536 & \textbf{0.0072} & 0.0629 & 0.0220 & 0.0947 & 0.0122 & 0.1011 & 0.0166 & 0.0711 \\
& 192 & \textbf{0.0078} & \textbf{0.0583} & \underline{0.0084} & \underline{0.0590} & 0.0096 & 0.0643 & 0.0471 & 0.1623 & 0.0135 & 0.0809 & 0.0256 & 0.0979 & 0.0137 & 0.1020 & 0.0271 & 0.0901 \\
& 336 & \textbf{0.0082} & \textbf{0.0579} & \underline{0.0096} & \underline{0.0624} & 0.0233 & 0.0687 & 0.0790 & 0.2717 & 0.0193 & 0.1570 & 0.0294 & 0.1057 & 0.0141 & 0.1097 & 0.0404 & 0.1274 \\
& 720 & \textbf{0.0105} & \textbf{0.0678} & \underline{0.0127} & \underline{0.0750} & 0.0148 & 0.0804 & 0.4471 & 0.4609 & 0.0289 & 0.1960 & 0.0417 & 0.1239 & 0.0231 & 0.1206 & 0.0654 & 0.1655 \\
\hline
\end{tabular}
\end{table*}

\begin{figure*}[!h] % Use !h to force placement directly after the table
    \centering
    \includegraphics[width=\textwidth, keepaspectratio]{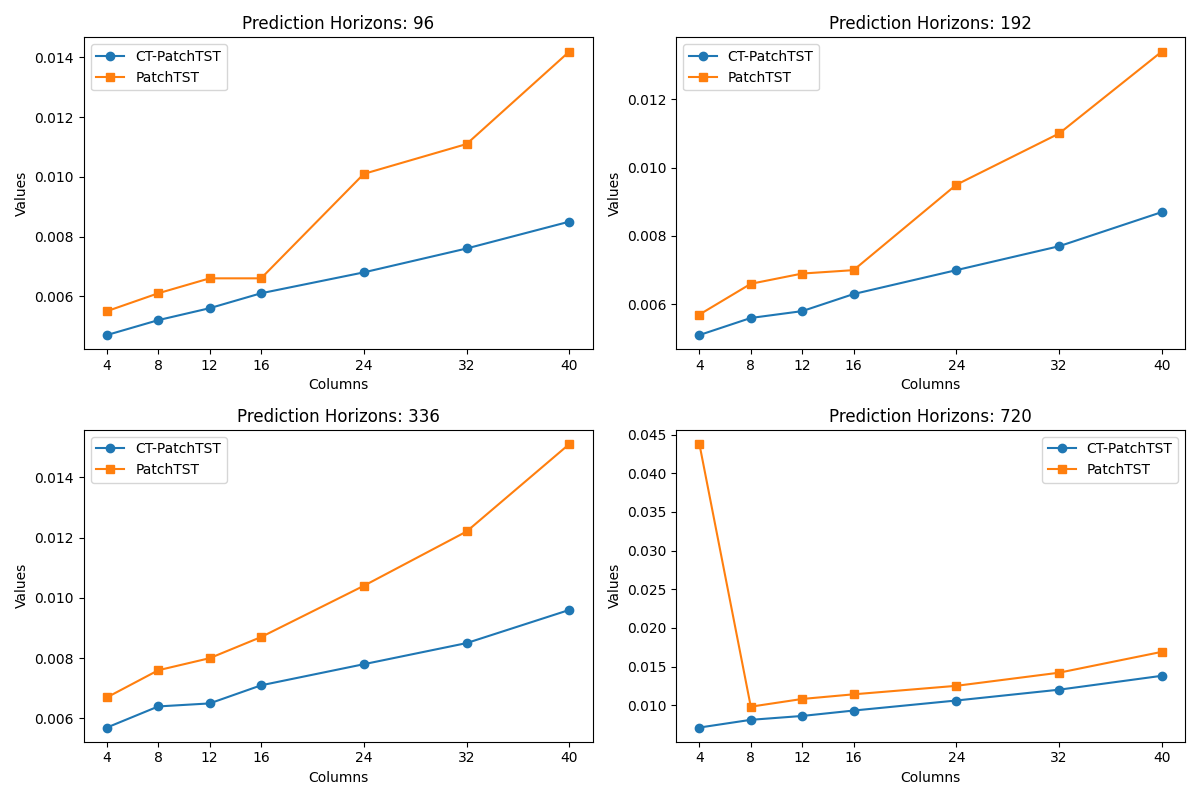}
    \caption{The forecasting performance (MSE) is evaluated on the dataset using CT-PatchTST and PatchTST models, with a fixed look-back window of 336 and forecasting lengths \( h \in \{96, 192, 336, 720\} \). The patch length \( P \) is varied across \(\{4, 8, 12, 16, 24, 32, 40\}\) to investigate its impact on model performance.}
    \label{fig:enter-label6}
\end{figure*}

% First figure (1 and 2)
\begin{figure*}
    \centering
    % Subfigure (a)
    \includegraphics[width=0.85\textwidth, keepaspectratio]{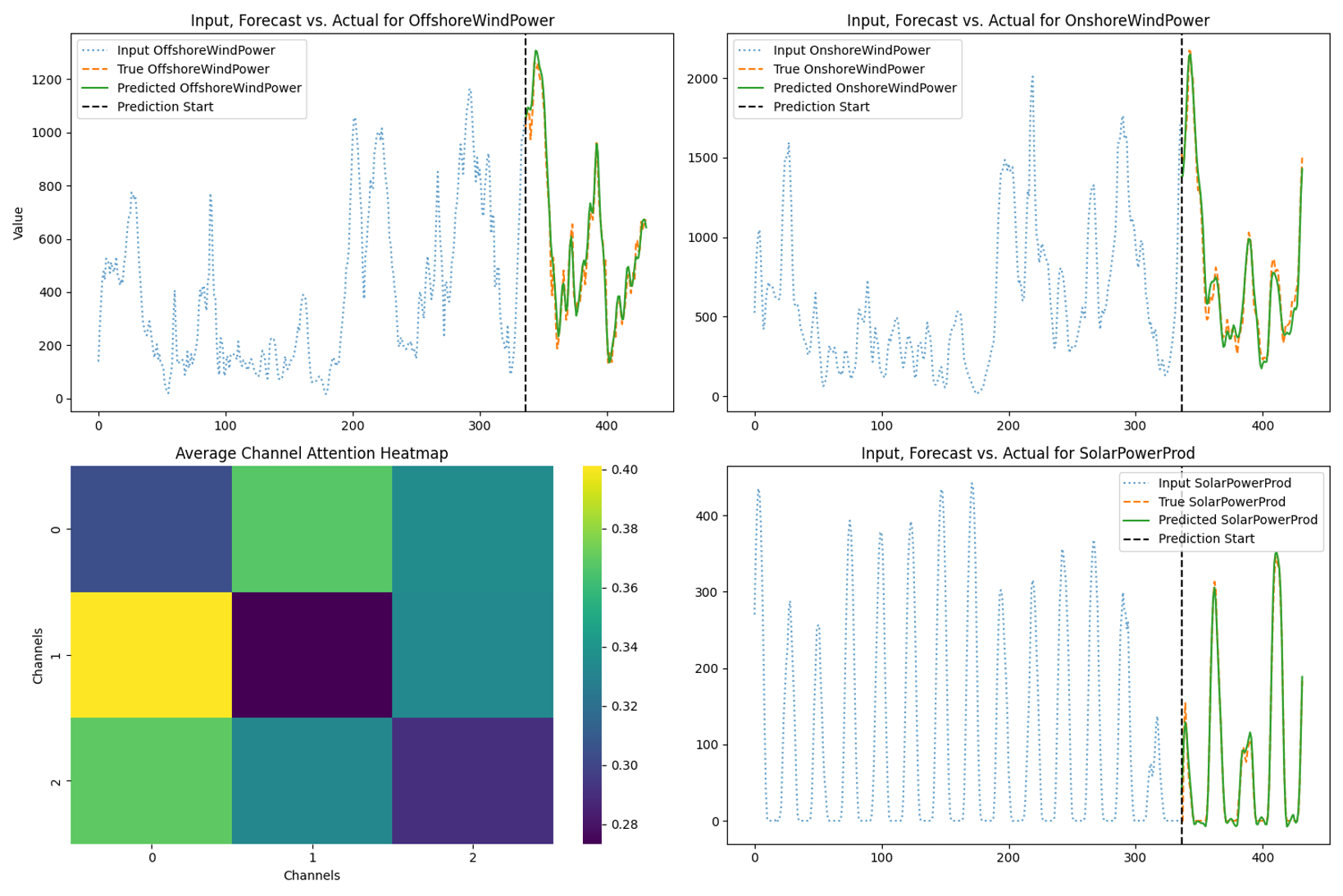}
    %\caption*{(a)} % Subcaption for (a)
    % Subfigure (b)
    \includegraphics[width=0.85\textwidth, keepaspectratio]{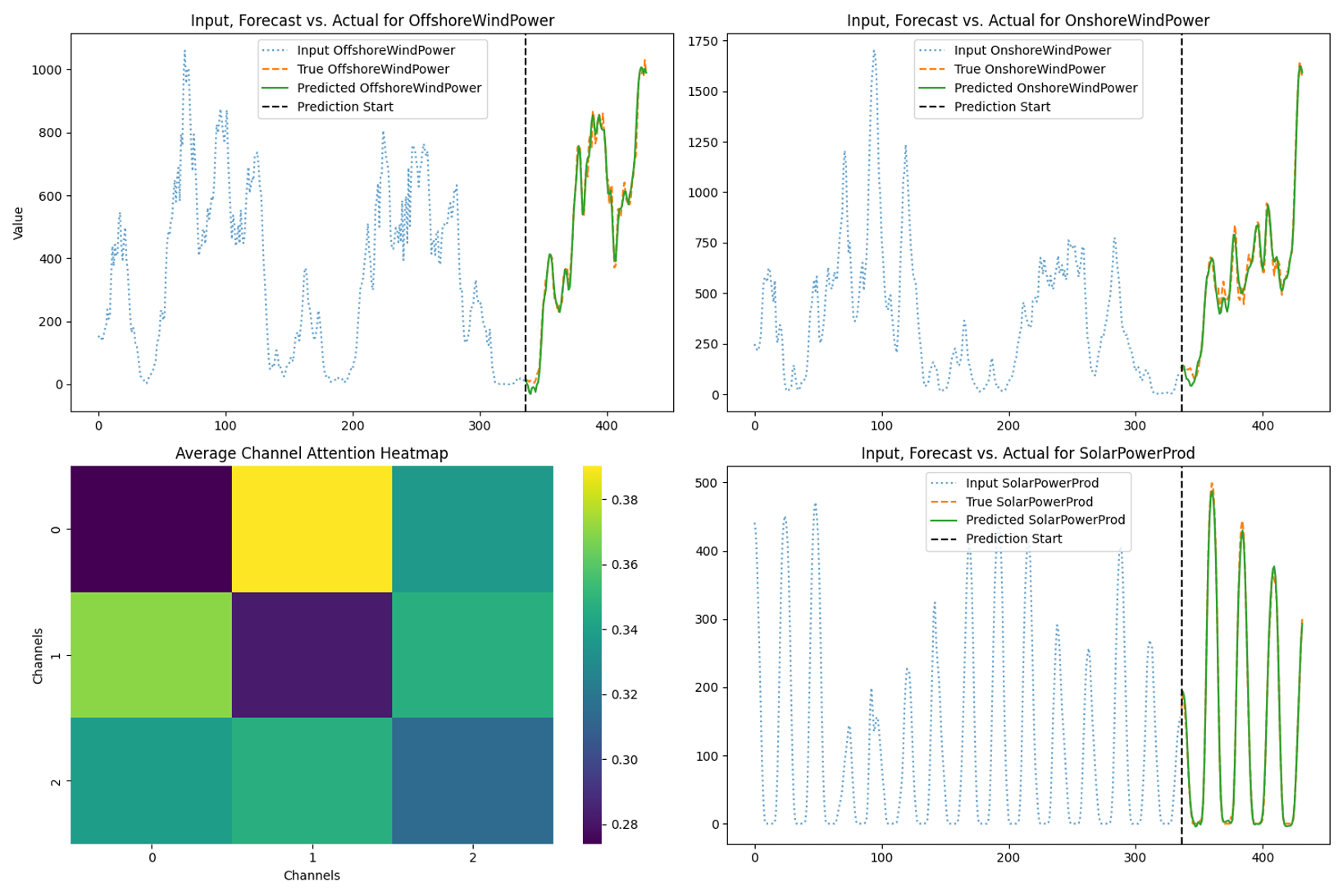}
    %\caption*{(b)} % Subcaption for (b)
    \label{fig:part7}
\end{figure*}

% Second figure (3 and 4)
\begin{figure*}
    \centering
    % Subfigure (c)
    \includegraphics[width=0.85\textwidth, keepaspectratio]{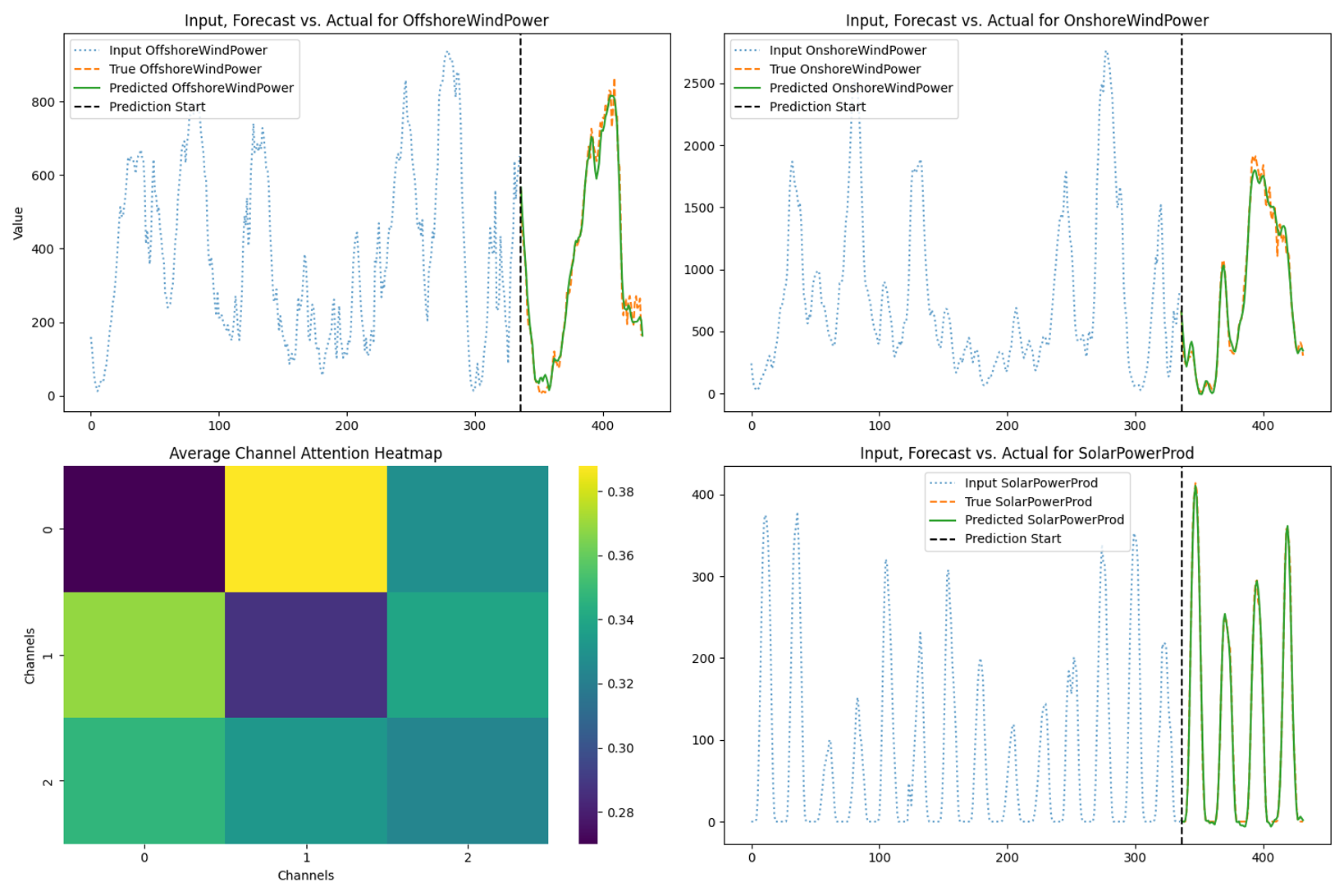}
    %\caption*{(c)} % Subcaption for (c)
    % Subfigure (d)
    \includegraphics[width=0.85\textwidth, keepaspectratio]{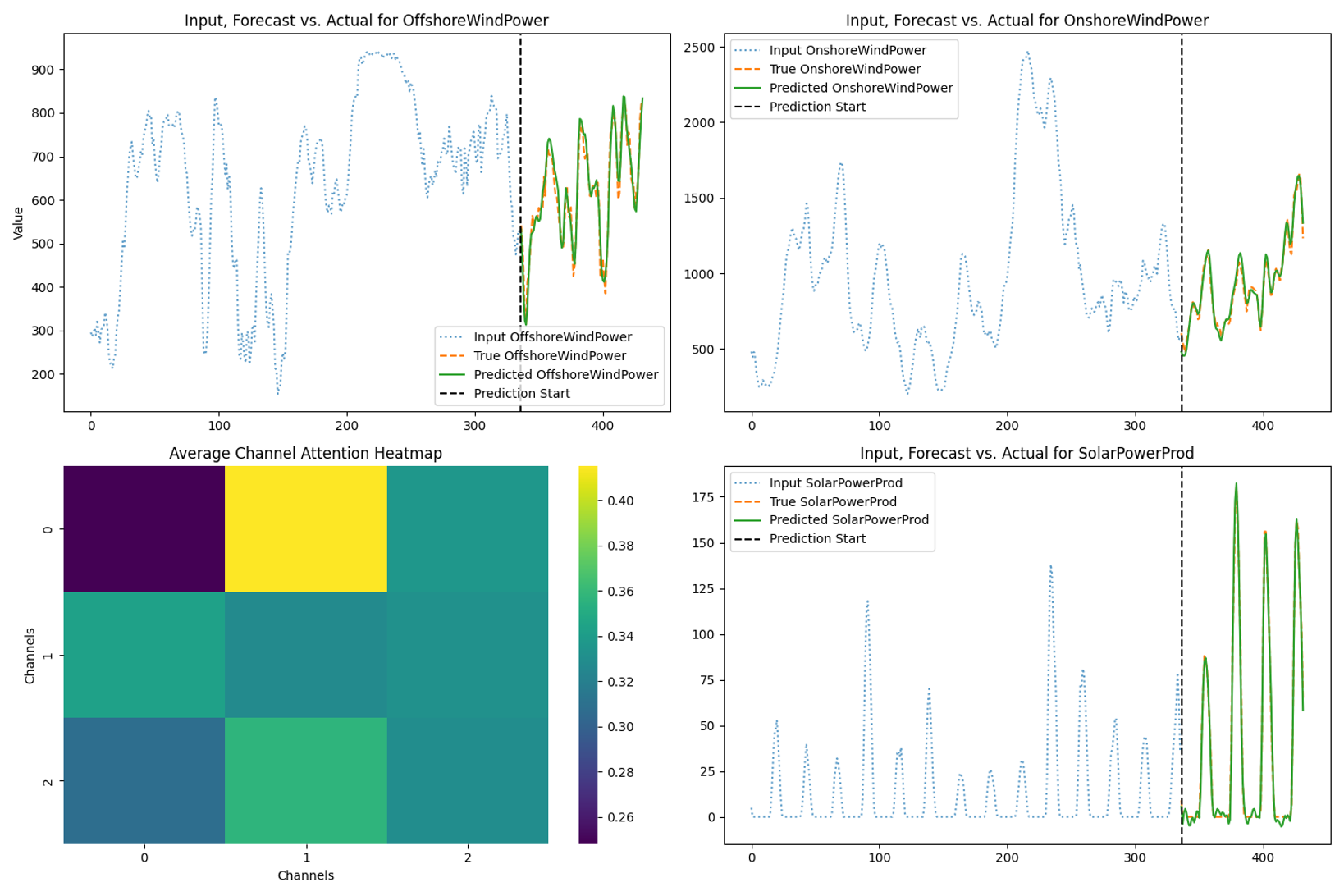}
    %\caption*{(d)} % Subcaption for (d)

    \caption{Attention maps of the Channel Attention mechanism and the forecasting results of selected time series, generated using CA-PatchTST. Experiments are conducted with a look-back window of 336 and a forecasting length of 96. Additionally, the average channel attention across all patches is computed to provide a comprehensive view of the attention distribution.}
    \label{fig:part8}
\end{figure*}

\subsection{Ablation Study}
In this section, we conducted two comprehensive ablation experiments to thoroughly examine the influence of varying patch lengths and prediction horizons on the performance of CT-PatchTST and to compare its effectiveness against the baseline PatchTST model with DK1 (Western Denmark) dataset. The first experiment focused on evaluating the impact of altering patch lengths, which were adjusted within a range from 4 to 40, to understand how the model’s ability to capture local and global patterns in multivariate time series data changes with different granularities. The second experiment explored the effect of varying prediction horizons, ranging from 96 to 720, to assess how well the models adapt to forecasting tasks of different temporal scales. 

The results of these experiments, as illustrated in Fig. 4, reveal that CT-PatchTST consistently outperforms PatchTST across all tested configurations. This consistent superiority demonstrates the robustness and adaptability of CT-PatchTST, particularly in capturing intricate inter-channel relationships and temporal dependencies within the data. The dual attention mechanisms, incorporating both channel and time dimensions, allow CT-PatchTST to extract more meaningful patterns compared to the baseline model. 

Moreover, these findings highlight the enhanced ability of CT-PatchTST to handle diverse forecasting scenarios, regardless of patch length or prediction horizon. Such versatility makes CT-PatchTST a powerful tool for multivariate time series forecasting, capable of addressing a wide range of real-world applications. By integrating more detailed insights into the interplay between channel and time dimensions, CT-PatchTST achieves significant improvements over the baseline PatchTST model, further validating its efficacy in complex forecasting tasks.

\subsection{Visualization Analysis}
To further validate the significance of channel attention in multivariate time series modeling, we visualized the average heat map generated by the channel attention mechanism. This visualization, along with the corresponding prediction results, is presented in Fig. 5, providing a comprehensive view of how channel attention operates within the model. The channel average attention map vividly demonstrates the ability of channel attention to capture and learn the intricate relationships among different channels, effectively highlighting the dependencies and interactions that exist within the multivariate data. 

By leveraging this mechanism, the newly generated sequence retains the essential information from the original channels while simultaneously incorporating valuable insights derived from other channels. This enriched representation not only ensures that the inter-channel relationships are preserved but also amplifies the relevance of the extracted features for subsequent tasks. As a result, the model becomes better equipped to account for inter-channel dependencies, ultimately leading to enhanced performance in forecasting applications.

These findings underscore the pivotal role of channel attention in multivariate time series modeling, particularly in scenarios where complex interactions between variables play a crucial role. By effectively exploiting these intricate interdependencies, channel attention enables the model to produce more accurate and reliable predictions. This highlights its potential as a key component in advancing the field of multivariate time series forecasting, paving the way for more robust and interpretable modeling techniques.

\section{Future Work}
Building on the current research, there is potential to further enhance the performance of multichannel time-series forecasting models. In the current implementation of channel attention, only the relationships between different channels within the same input patch are considered. However, this approach may overlook the influence of interactions between different input patches on different channels. To address this limitation, future work could explore strategies for better integrating channel and temporal information simultaneously. By developing a more holistic approach that captures both inter-channel and inter-patch dependencies, the model's efficiency and forecasting accuracy can be further improved, paving the way for more robust solutions in multichannel time-series forecasting.

\section{Conclusion}
In this paper, we propose an enhanced version of the PatchTST model, CT-PatchTST, which simultaneously considers both channel and temporal dimensions in multivariate time series forecasting. By addressing a key limitation of the original PatchTST, CT-PatchTST incorporates channel attention to effectively capture inter-channel relationships and time attention to model temporal dependencies, enabling a more comprehensive understanding of the complex patterns in multivariate time series data. Experimental results demonstrate that CT-PatchTST consistently outperforms other state-of-the-art models across various settings, highlighting its superior ability to handle diverse forecasting scenarios. Additionally, a detailed comparison of CT-PatchTST and the original PatchTST under varying patch lengths and forecast horizons further validates the effectiveness and adaptability of our approach. Looking ahead, we aim to explore more advanced techniques for jointly modeling the interrelationship between channels and temporal dynamics, with the goal of enhancing the predictive accuracy and robustness of CT-PatchTST, thereby contributing to advancements in multivariate time series forecasting.

\bibliographystyle{IEEEtran}
\bibliography{report}

\end{document}